\documentclass[a4paper,11pt]{article}

\usepackage{epsfig}
\usepackage{amsfonts}
\usepackage{times}
\usepackage{graphicx}

\usepackage{epsfig}
\usepackage{url}
\usepackage{array}
\usepackage{amsfonts}
\usepackage{amsmath}
\usepackage{amssymb}
\usepackage{latexsym}
\usepackage{float}
\usepackage{multirow}

\newcommand{\ben}{\begin{enumerate}}
\newcommand{\een}{\end{enumerate}}
\newcommand{\bi}{\begin{itemize}}
\newcommand{\ei}{\end{itemize}}
\newcommand{\be}{\begin{equation}}
\newcommand{\ee}{\end{equation}}
\newcommand{\bea}{\begin{eqnarray}}
\newcommand{\eea}{\end{eqnarray}}
\newcommand{\ba}{\begin{array}}
\newcommand{\ea}{\end{array}}
\newcommand{\bc}{\begin{center}}
\newcommand{\ec}{\end{center}}
\newcommand{\bt}{\begin{tabular}}
\newcommand{\et}{\end{tabular}}
\newcommand{\bfig}{\begin{figure}[htb]}
\newcommand{\efig}{\end{figure}}

\newcommand{\V}{\mathbf{V}}
\newcommand{\Vh}{\hat{\mathbf{V}}}
\newcommand{\W}{\mathbf{W}}
\renewcommand{\H}{\mathbf{H}}
\newcommand{\Rbb}{\mathbb{R}}

\begin{document}

%\dsp
% paper title
\title{Efficient Bayesian Community Detection using Non-negative Matrix Factorisation}

\author{
Ioannis Psorakis \thanks{Authors in alphabetic order.}~\thanks{Pattern
Analysis \& Machine Learning Research Group, Department of Engineering
Science, University of     Oxford. Parks Road, Oxford, OX1 3PJ
U.K. Email     yannis@robots.ox.ac.uk}~~, Stephen Roberts \thanks{Pattern
Analysis \& Machine Learning  Research Group. Email sjrob@robots.ox.ac.uk}~~ and Ben Sheldon \thanks{Dept
Zoology, University of Oxford}
}

% make the title area
\maketitle

\begin{abstract}
Identifying overlapping communities in networks is a challenging
task. In this work we present a novel approach to community detection
that utilises the Bayesian non-negative matrix factorisation (NMF) model to produce a probabilistic output
for node memberships. The scheme has the advantage of computational efficiency, soft community membership and an intuitive foundation. We present the performance of the method
against a variety of benchmark problems and compare and contrast it to several other algorithms for community detection. Our approach performs favourably compared to other methods at a fraction of the computational costs.
\end{abstract}

\section{Introduction}
\label{sec:Introduction}

The network paradigm is widely used to model real-world complex systems, by focusing on the pattern of associations between their structural components. A system is captured as a mathematical graph, where nodes (or vertices) denote the presence of an entity and edges (or links) signify some sort of association (or interaction). In contrast to other data manipulation approaches where each element is described by a set of attributes (for example $\mathbf{x} \in \mathbb{R}^D$), here our data is captured in a \emph{relational} form and inferences are made primarily based on their connectivity patterns.

Real-world networks differ from the classic Erdos-Renyi random graph because the presence of a link between two nodes is not generated by a Bernoulli trial with same success probability across all possible pairs. Instead, real-world networks exhibit an inhomogeneous distribution of edges among vertices \cite{fortunato_overview}, creating `hotspots' of hightened connectivity. These \emph{modules} or \emph{communities} are densely connected, relatively independent compartments \cite{fortunato_overview} \cite{LaFort} of the network that account for its form and function as a system \cite{NewmanBook}. The intuition behind that \emph{mesoscopic} organisation of networks is intuitively straightforward, with many examples from everyday life; human social networks consist of cliques of friends, Web pages can be grouped into collections with similar topic, etc. 

\begin{figure}[htbp]
   \centering
   \includegraphics[angle=0,scale=0.6]{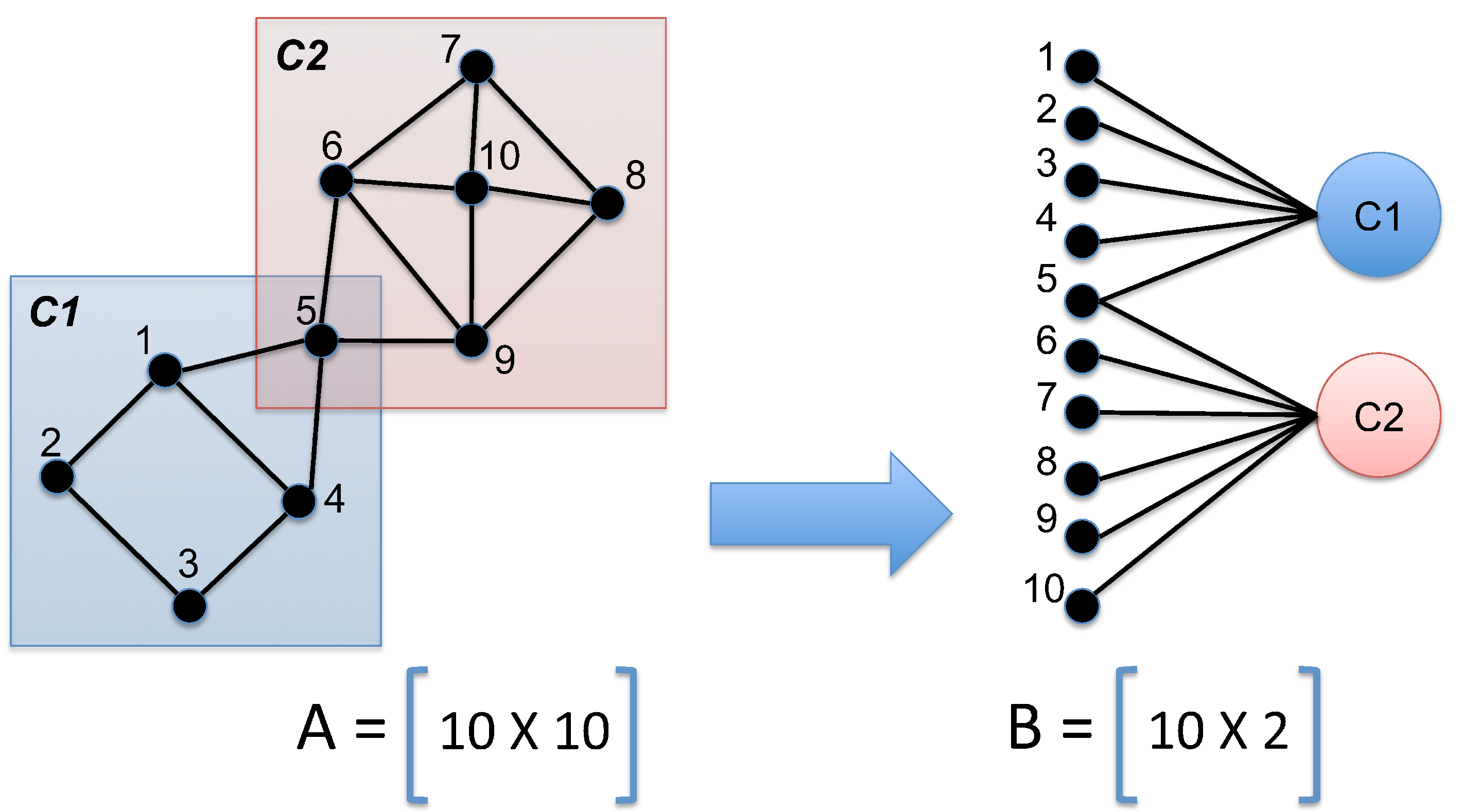}
   \caption{\label{small_network} The problem of community detection can be viewed as mapping our network to a bipartite graph, where the first class corresponds to individual nodes and the second to communities. Links connecting nodes to multiple communities help us capture \emph{overlapping} phenomena, in which an individual can participate in many groups.}
\end{figure}

As an example, consider the simple undirected graph of Fig. \ref{small_network}, described by an adjacency matrix $\mathbf{A} \in \mathbb{R}^{N \times N}$ where we can immediately distinguish the two densely connected compartments C1 and C2. As we are not sure of the membership of node 5, it is fairly reasonable to consider it as an \emph{overlap} of C1 and C2. Therefore, we can express our community partition as an expansion of our network to a bipartite graph with incidence matrix $\mathbf{B} \in \mathbb{R}^{N \times K}$ so that $b_{ik} = 1$ denotes that node $i$ belongs to a group $k$ and is zero otherwise. Although the human eye is an excellent analytic tool for simple visualised data \cite{newman_overview},  the algorithmic process of identifying the number of groups, classifying their members and spotting overlaps in any given network is far from straightforward.

Extracting community structure from a network is a considerably challenging task \cite{NewmanBook}, both as an inference and as a computational efficiency problem. One of the main reasons is that there is no formal, application-independent definition of what a community actually is \cite{fortunato_overview}; we simply accept that communities are node subsets with a `statistically surprising' link density, usually measured `modularity' $Q$ from Mark Newman and Michelle Girvan \cite{newman_girvan}. Nevertheless, this definition lies in the heart of modern community detection algorithms, manifested either as an exploration of configurations (such as $\mathbf{B}$ described above) that seek to approximately maximise $Q$ (as direct optimisation is NP-hard \cite{fortunato_overview}), or other techniques that exploit characteristics of the network topology to group together nodes with high mutual connectivity (and use $Q$ as a performance measure). A fairly comprehensive review of existing methods is provided in \cite{mason_overview} and \cite{fortunato_overview}, while \cite{danon} provides a table summarising the computational complexity of popular algorithms.

%As mentioned previously, a community detection algorithm must have both adequate module recognition and computational efficiency wise. Some of the most popular methods such as Newman-Girvan \cite{newman-girvan}, Extremal Optimisation \cite{eo} or Spectral Partitioning \cite{sp} produce good results in identifying communities but omit the \emph{overlapping} nature of communities. On the other hand, `soft-partitioning' methods such as Clique Percolation \cite{clique} are computationally intensive
The most significant drawbacks of modern community detection algorithms are \emph{hard partitioning} or/and \emph{computational complexity}. Many widely popular methods such as Girvan-Newman \cite{newman_girvan}, Extremal Optimisation \cite{eo} or Spectral Partitioning \cite{newman_sp} cannot account for the \emph{overlapping} nature of communities, which is an important characteristic of complex systems \cite{fortunato_overview}. On the other hand, the Clique Percolation Method (CPM) \cite{CPM} allows node assignment to multiple modules but does not provide some `degree of belief' on how strongly an individual belongs to a certain group.
%while other methods require multiple runs by introducing small pertrubations to the original network. 
Therefore, our aim is not only to model group overlaps but also capture the \emph{likelihood} of node memberships in a disciplined way, expressing each row of the incidence matrix $\mathbf{B}$ as a \emph{membership distribution} over communities. Additionally, we want to avoid the computational complexity issues of traditional combinatorial methods.

% =============================================================
%Steve's part
% In recent years there has been an increasing foundation of research in
% community detection and social network analysis. Much of this work is
% based around greedy allocation and efficient, energy based, algorithms
% for finding communities and allocating members.

Towards these goals, community detection can be seen as a
\textit{generative model} in a probabilistic framework.
This has the advantage that, in principle, fully Bayesian models may
be formulated in which \textit{priors} exist over all the model
parameters. This enables, for example, model selection (to determine
e.g. how many communities there are) and the principled handling of
uncertainty, noise and missing data. 

In this paper we consider the case in which constraints exist in our
beliefs regarding the generative process for the observed data. In particular
we investigate the issue of enforcing \textit{positivity} onto the
model. As
the model we consider is evaluated in a Bayesian manner, model selection
may be applied to infer the complexity of representation in the
solution space of inferred communities.

This paper is organised as follows. We first introduce the basic
concepts of the theory and our data decomposition goals.
Details of the Bayesian paradigm under which model
inference is performed are then presented. Representative results are
given in the next sections followed by conclusions and discussion.

\section{Methods} \label{sec:Methods}
We consider a matrix of observed interactions between a set of
$N$ \textit{atoms}, which in the context of community detection we consider
to be individuals. We consider an \textit{interaction matrix} denoted
$\V \in \Rbb_{+}^{N \times N}$ such that $v_{ij}$ represents a count
process detailing the number of interactions between atoms
(individuals) $i$ and $j$.
\footnote{A brief note on nomenclature - as the method described here has its
roots in the work in \textit{non-negative matrix factorization} (NMF),
we keep to the historic nomenclature of the latter to make easier
access for the reader who wishes to follow up primary references.}

\subsection{Decomposition} \label{sec:decomposition}
We consider the decomposition of the observed interaction matrix $\V$ as a linear
combination of $K$ canonical communities, each of which can be seen as a
latent (hidden) generator of interactions between atoms. Hence,
\be
\hat{v}_{ij} = \sum_{k=1}^{K} w_{ik} h_{kj} ,
\label{eq:linear_sum}
\ee
in which we regard the $w_{ik}$ as mixing coefficients and the $h_{kj}$ as elements forming a basis set of community structures.
The above equation may be re-written in matrix form by defining $\W \in
\Rbb^{N \times K}$ and $\H \in \Rbb^{K \times N}$
\be
\Vh = \W \H .
\label{eq:WH}
\ee
Without constraint, Equation \ref{eq:WH} is ill-posed, i.e. an infinite
number of equivalent solutions exist. Many well-known decomposition methods can be seen as
members of a family of approaches which impose constraints in the
solution space to make the matrix-product decomposition well-posed.

\paragraph{PCA:} Principal Component Analysis
\cite{Jolliffe86Principal} avoids the problem of an ill-posed solution
space by making several constraints. The first is that the observed data distrbution and the basis
are Gaussian distributed (in that only second-order
statistics are employed in the PCA formalism) and that the
basis is orthogonal. This still leaves a permutation degree of
freedom, which is removed by sorting the basis in order of variance
(in effect an ordering of the eigenvalues).
\paragraph{ICA:} Independent Component Analysis \cite{Comon:94,ICA_book:01} makes the solutions to Equation \ref{eq:WH}
well-posed by forcing statistical independence between the components of the
basis without making strong assumptions regarding the Gaussianity of the component distributions, unlike PCA. This gives rise to a methodology which allows projective decompositions similar to PCA for non-Gaussian data.
\paragraph{NMF:} Non-negative Matrix Factorization makes the
assumption that all the elements of matrices $\V , \W, \H$ lie in $\Rbb_{+}$. This is
enough, up to an arbitrary scaling degree of freedom, to ensure that
solutions to Equation \ref{eq:WH} are well-posed.
The latter assumptions of non-negativity  match our prior beliefs regarding the
generation of the observed interaction count data, $\V$, and we extend our discussion of
NMF in the following sections.

\subsection{Generative Model} \label{sec:gen_model}
We consider the observed data to be modelled as a Poisson process with
expectations given by the elements of $\Vh = \W \H$. This Poisson model lies at the core of the NMF methodology
first developed by Lee \& Seung \cite{Lee+Seung:99}. The standard 
maximum-likelihood solution is to find $\W, \H$ such that $p(\V |
\W,\H)$ is maximized, or alternately, the energy function $-\log p(\V |
\W,\H)$ is minimized.

Consider an element $v$ of $\V$ and the associated expected counts from the model, $\hat{v}$. The negative log likelihood of $v$ under a Poisson model is:
\begin{equation} \label{eq:log_like1}
	-\log p(v|\hat{v}) = -v \log \hat{v} + \hat{v} + \log v!.
\end{equation}
Using the Stirling approximation to second order, namely
\begin{equation} \label{eq:stirling}
 \log v! = v \log v - v + \frac{1}{2} \log (2\pi v),
\end{equation}
so Equation \ref{eq:log_like1} can be written as,
\begin{equation} \label{eq:log_like2}
	-\log p(v|\hat{v}) = v \log \left( \frac{v}{\hat{v}} \right ) + \hat{v} - v + \frac{1}{2} \log (2\pi v)
\end{equation}
and the full negative log-likelihood for all the observed data as
\begin{equation} \label{eq:log_like3}
 -\log p(\V | \hat{\V}) = -\sum_{i} \sum_{j} \log p(v_{ij} | \hat{v}_{ij}).
\end{equation}

\subsubsection{Shrinkage hyperparameters} \label{sec:betas}
As each of the $k \in \{1...K\}$ columns of $\W$ and rows of $\H$ represents the
contribution from a single latent community (as per Equation
\ref{eq:linear_sum}) we allow for different shrinkage hyperparameters, defined as the set of $\{ \beta_k \}$. Following the development of this model in \cite{tan09} and similar models for probabilistic PCA \cite{Tipping:99} and ICA \cite{Choudrey+Roberts:03,Roberts+Choudrey:05} we place independent half-normal priors over the columns of $\W$ and rows of $\H$ in which the $\beta_k$ may be seen as precision (inverse variance) parameters:
\begin{eqnarray} \label{eq:HN1}
 p(w_{ik} | \beta_k) = \mathcal{HN}(w_{ik} | 0, \beta_k^{-1}) \nonumber \\
 p(h_{kj} | \beta_k) = \mathcal{HN}(h_{kj} | 0, \beta_k^{-1}),
\end{eqnarray}
where
\begin{equation} \label{eq:HN2}
 \mathcal{HN}(x | 0, \beta^{-1}) = \sqrt{\frac{2}{\pi}} \beta^{1/2} \exp \left (-\frac{1}{2}\beta x^2 \right ).
\end{equation}
Defining the vector $\boldsymbol{\beta}$ as $[\beta_1, ..., \beta_K]$, this leads to negative log priors over $\W$ and $\H$ as:
\begin{eqnarray}
 -\log p(\W|\boldsymbol{\beta}) = \sum_i \sum_k \frac{1}{2}\beta_k w_{ik}^2 - \frac{F}{2} \log \beta_k + \rm{const}, \nonumber \\
 -\log p(\H|\boldsymbol{\beta}) = \sum_k \sum_j \frac{1}{2}\beta_k h_{kj}^2 - \frac{N}{2} \log \beta_k + \rm{const}.
 \label{eq:log_priors}
\end{eqnarray}
The net effect of $\boldsymbol{\beta}$ on the elements of $\W$ and $\H$ may be considered as follows. As the negative log probability may be regarded as an error or energy function our goal is to descend its surface to a point of minimum. Consider the negative derivative w.r.t. a single element of $\W$ or $\H$ of the above Equations, e.g.
\begin{equation}
 - \frac{\partial \left (-\log p(\W|\boldsymbol{\beta}) \right )}{\partial w_{ik}} = - \beta_k w_{ik}.
\end{equation}
Incremental changes in $w_{ik}$, as we iterate towards a solution, are proportional to the negative gradient of the energy function, so the effect of the prior is to promote a \textit{shrinkage} to zero of $w_{ik}$ with a rate constant proportional to $\beta_k$. A large $\beta_k$ represents a belief that the half-normal distribution over $w_{ik}$ has small variance, and hence $w_{ik}$ is expected to lie close to zero. As we shall see, the priors and the likelihood function (quantifying how well we explain the data) are combined with the net effect that columns of $\W$ (and rows of $\H$) which have little effect in changing how well we explain the observed data will shrink close to zero. This generic approach is well known in the statistics literature, as \textit{shrinkage} or \textit{ridge regression} \cite{Bernardo+Smith:94} and in the machine learning community as \textit{automatic relevance determination} \cite{bishop2007}.

Finally we must place prior distributions over the $\beta_k$. We assume the set of $\beta_k$ are independent \footnote{This corresponds to the belief that the existence of one community is not dependent upon others. Clearly, there will be situations in which this can be extended to allow for a full inter-dependency between communities. We do not consider this here, however. Allowing dependency is similar to the notion of \textit{structure priors} discussed in \cite{Penny_Roberts_MAR:02}.} and as these are scale hyperparameters we place a standard Gamma distribution over them \cite{Bernardo+Smith:94}:
\begin{equation} \label{eq:gammadist}
 p(\beta_k | a_k, b_k) = \frac{b_k^{a_k}}{\Gamma (a_k)} \beta_k^{a_k-1} \exp \left (-\beta_k b_k \right ),
\end{equation}
in which the hyper-hyperparameters $a_k, b_k$ defining the gamma distrubution over $\beta_k$ are fixed. The negative log of the probability distribution over $\boldsymbol{\beta}$ is hence,
\begin{equation} \label{eq:pbeta}
 - \log p(\boldsymbol{\beta}) = \sum_k \left [ \beta_k b_k - (a_k-1) \log \beta_k \right ] + \rm{const}.
\end{equation}

\subsubsection{Overall posterior cost function} \label{sec:map}
\begin{figure}[htbp]
   \centering
   \includegraphics[angle=0,scale=0.3]{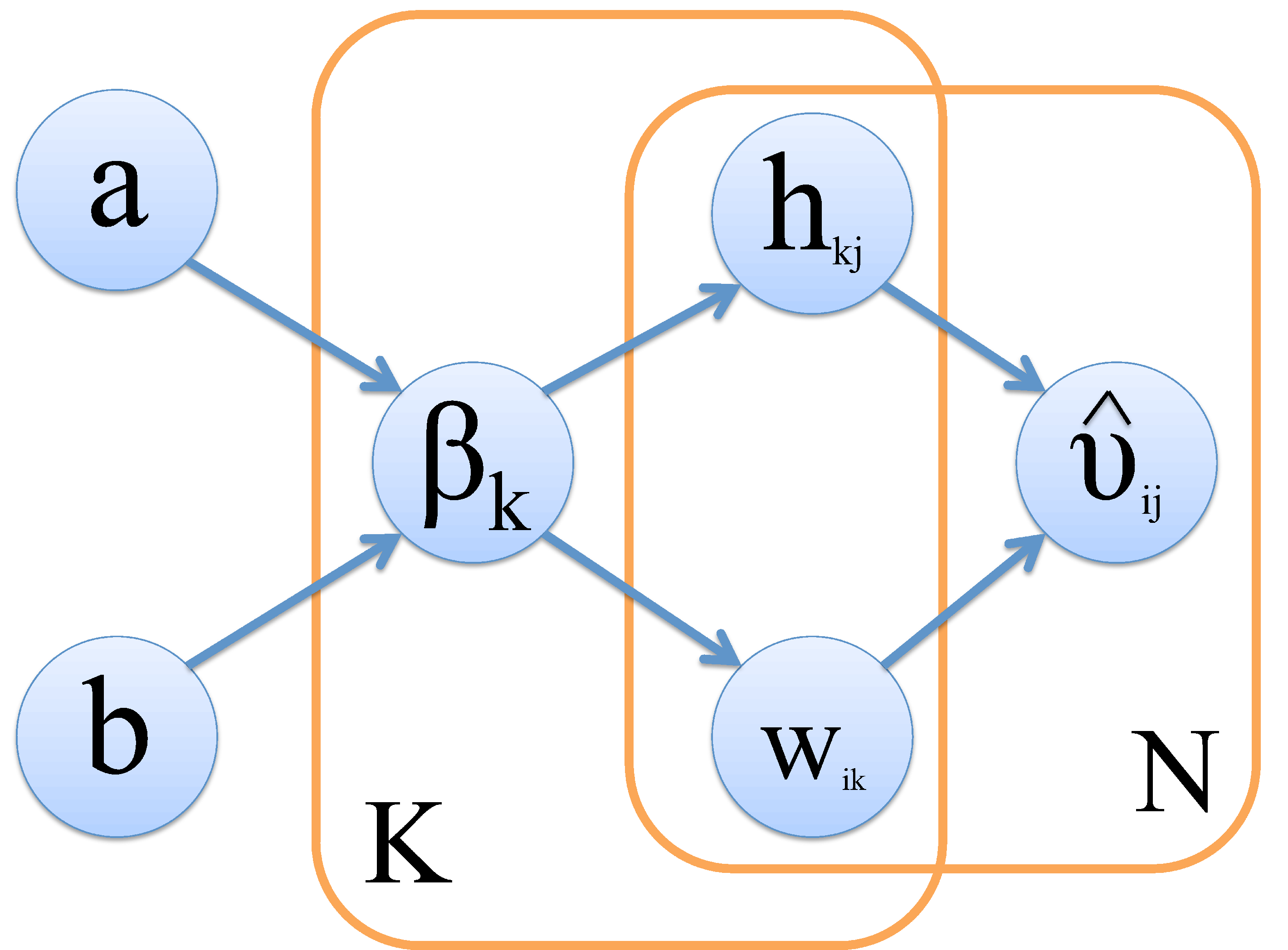}
   \caption{\label{fig:model}Graphical model showing the generation of count processes, $\V$ from the latent structure $\W$ and $\H$ the components of which have scale hyperparameters $\beta$. The hyper-hyperparameters $a,b$ are fixed in the model.}
\end{figure}

Figure \ref{fig:model} shows the generative graphical model for the NMF method.
The joint distribution over all variables is
\begin{equation} \label{eq:joint}
 p(\V,\W,\H,\boldsymbol{\beta}) = p(\V|\W,\H)p(\W|\boldsymbol{\beta})p(\H|\boldsymbol{\beta})p(\boldsymbol{\beta}),
\end{equation}
and the model posterior over all parameters, given the observations is:
\begin{equation}
 p(\W,\H,\boldsymbol{\beta} | \V) = \frac{p(\V,\W,\H,\boldsymbol{\beta})}{p(\V)}.
\end{equation}
Noting that $p(\V)$ is a constant w.r.t. the inference over the model's free parameters, we aim hence to maximize the model posterior given the observations. This is equivalent to minimizing the negative log posterior, which we may regard as an energy (or error) function, $\mathcal{U}$ say. We hence define
\begin{equation} \label{eq:U}
 \mathcal{U} = \log p(\V|\W,\H) - \log p(\W|\boldsymbol{\beta}) - \log p(\H|\boldsymbol{\beta}) - \log p(\boldsymbol{\beta}).
\end{equation}
Expanding this expression using the results from Equations \ref{eq:log_like2}, \ref{eq:log_like3}, \ref{eq:log_priors} and \ref{eq:pbeta} and collating all terms independent of the model parameters into a constant, gives:
\begin{eqnarray} \label{eq:post}
 \mathcal{U} & = & \sum_i \sum_j \left [ v_{ij} \log \left ( \frac{v_{ij}}{\hat{v}_{ij}} \right ) + \hat{v}_{ij} \right ] \nonumber \\
& + & \frac{1}{2} \sum_{k} \left [ \left (\sum_f \beta_k w_{ik}^2 \right ) + \left (\sum_n \beta_k h_{kj}^2 \right ) - (F+N) \log \beta_k \right ] \nonumber \\
& + & \sum_k \left [ \beta_k b_k - (a_k-1) \log \beta_k \right ] + \rm{const}.
\end{eqnarray}

\subsection{Parameter Inference} \label{sec:inference}
There are a variety of approaches one could take to infer $\W,\H,\boldsymbol{\beta}$ given Equation \ref{eq:post}. In this paper we follow \cite{Lee+Seung:99,Lee00algorithmsfor,Berry06algorithmsand,tan09} and utilize a rapid fixed point maximum a posteriori (MAP) algorithm which guarantees to preserve the non-negativity of all parameters in the model. At each iteration the following re-estimations are made,
\begin{eqnarray}
 \H & \leftarrow & \left ( \frac{\H}{\W^{\sf T} \mathbf{1} + \mathbf{B}\H} \right ) \W^{\sf T} \left ( \frac{\V}{\W\H}  \right ) \label{eq:Hupdate}\\
 \W & \leftarrow & \left ( \frac{\W}{\mathbf{1}\H^{\sf T} + \W \mathbf{B}} \right ) \left ( \frac{\V}{\W\H} \right ) \H^{\sf T}
 \label{eq:Wupdate}
\end{eqnarray}
in which we define $\mathbf{B}$ as having the elements $\beta_k$ along its diagonal and zeros elsewhere and $\mathbf{1}$ is a vector of ones. We keep to the notation convention that $\left ( \frac{\mathbf{X}}{\mathbf{Y}} \right )$ represents \textit{element-by-element} division and does \textit{not} represent $\mathbf{XY}^{-1}$. The values of $\beta_k$ are re-estimated by setting to zero the derivative w.r.t. the $\beta_k$ of the energy function in Equation \ref{eq:post}. This gives an estimate for $\beta_k$ as 
\begin{equation} \label{eq:betaupdate}
 \beta_k \leftarrow \frac{ N +a_k-1 }{\frac{1}{2} \left (\sum_f w_{ik}^2 + \sum_n h_{kj}^2 \right ) + b_k }
\end{equation}
which is the same as the update equation detailed in \cite{tan09}. The algorithm proceeds by cycling through Equations \ref{eq:Hupdate},\ref{eq:Wupdate}, \ref{eq:betaupdate} until a convergence criterion or maximum number of iterations is reached. We note that a fully Bayesian approach to NMF is developed using variational inference in \cite{Cemgil:09}. This offers certain potential improvements over the maximum a posteriori (MAP) solution at the expense of computational speed. The latter we regard as a very important feature of any algorithm for community detection and the current emphasis of our work is directed by computational efficiency.

\subsection{Probabilistic community membership} \label{sec:fuzzMem}
As we may write the observed data as a linear combination of community basis structures and the mixing fractions are strictly non-negative, the model is identical to a \textit{mixture model} in which the elements $w_{ik}$ denote the relative importance of community $k$ in explaining the observed interactions associated with member $i$. Under the assumption that the $i$-th member's interactions are explained by some community memberships, it is reasonable to define degrees of community membership, $\pi_{ik}$, which sum to unity for each member, as:
\begin{equation} \label{eq:fuzzMem}
 \pi_{ik} = \frac{w_{ik}}{\sum_{k'} w_{ik'}}.
\end{equation}
A greedy community allocation scheme for member $i$ is easily achieved, if desired, by choosing the community $k*$ which is the argmax of either the $w_{ik}$ or $\pi_{ik}$.

%\subsection{Missing Data} \label{sec:missing_data}

\section{Results} \label{sec:results}

In this section we demonstrate the performance of NMF-based community detection against a variety of benchmark problems. We start with a toy network to illustrate the intuition behind our clustering methodology using graphical examples. Afterwards, we continue by testing our method against artificial problems with observed community structure. Finally, we test our method against popular real-world networks of various sizes and levels of community cohesiveness.

\subsection{Initialization}
In all the results presented in this paper we allow the maximum number of possible communities $K$ to equal $N$, the number of members. The hyper-hyperparameters, $a$ and $b$, which govern the scale of the shrinkage hyperparameters, $\beta_k$, are fixed at $a =1, b = 2$ so that $\beta_k$ all have vague distributions over them. The initial matrices $\W, \H$ each have elements drawn independently at random from a uniform distribution in the interval $[0,1]$.

The interaction matrix $\V$ we use for NMF is derived from the (weighted) adjacency matrix of each network with diagonal elements the strengths of each node (the sum of each row or column of the adjacency matrix).

\subsection{An Illustrative Example} \label{sec:An Illustrative Example}
\begin{figure}[htbp]
   \centering
   \includegraphics[angle=0,scale=0.6]{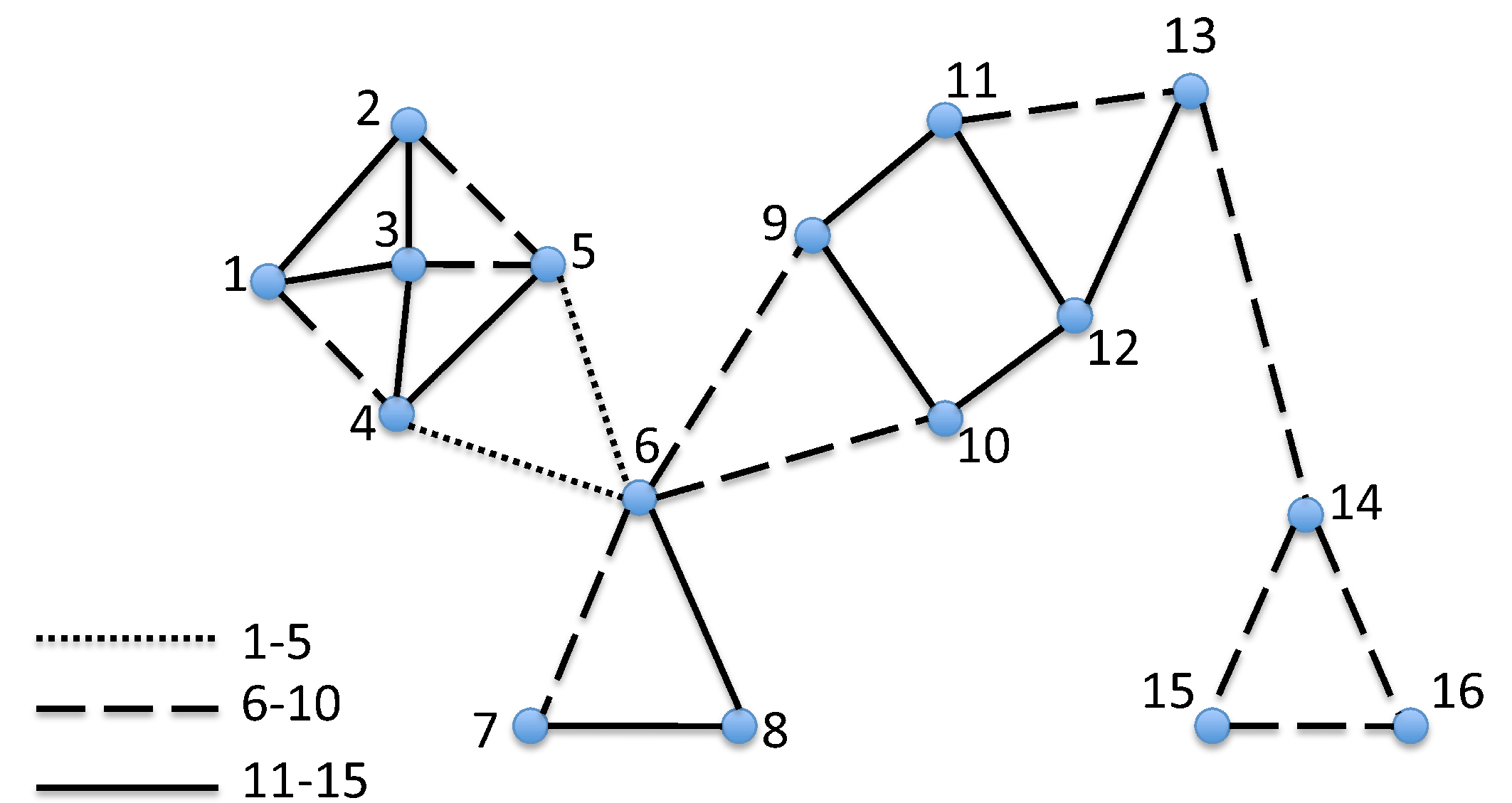}
   \caption{\label{toygraph} An undirected weighted toy graph with 16 nodes. Each pair of nodes has a different interaction strength as denoted by the different lines.}
\end{figure}

Consider the simple toy graph of Fig. \ref{toygraph} with $N=16$ nodes and $M=25$ edges of varying weights. We extract the mesoscopic (community) structure of this network using NMF, along with the popular Extremal Optimisation (EO) \cite{eo}, Spectral Partitioning (SP) \cite{newman_sp} and Weighted Clique Percolation Method (wCPM) \cite{wCPM}. 

\begin{figure}[htbp]
   \centering
   \includegraphics[angle=0,scale=0.6]{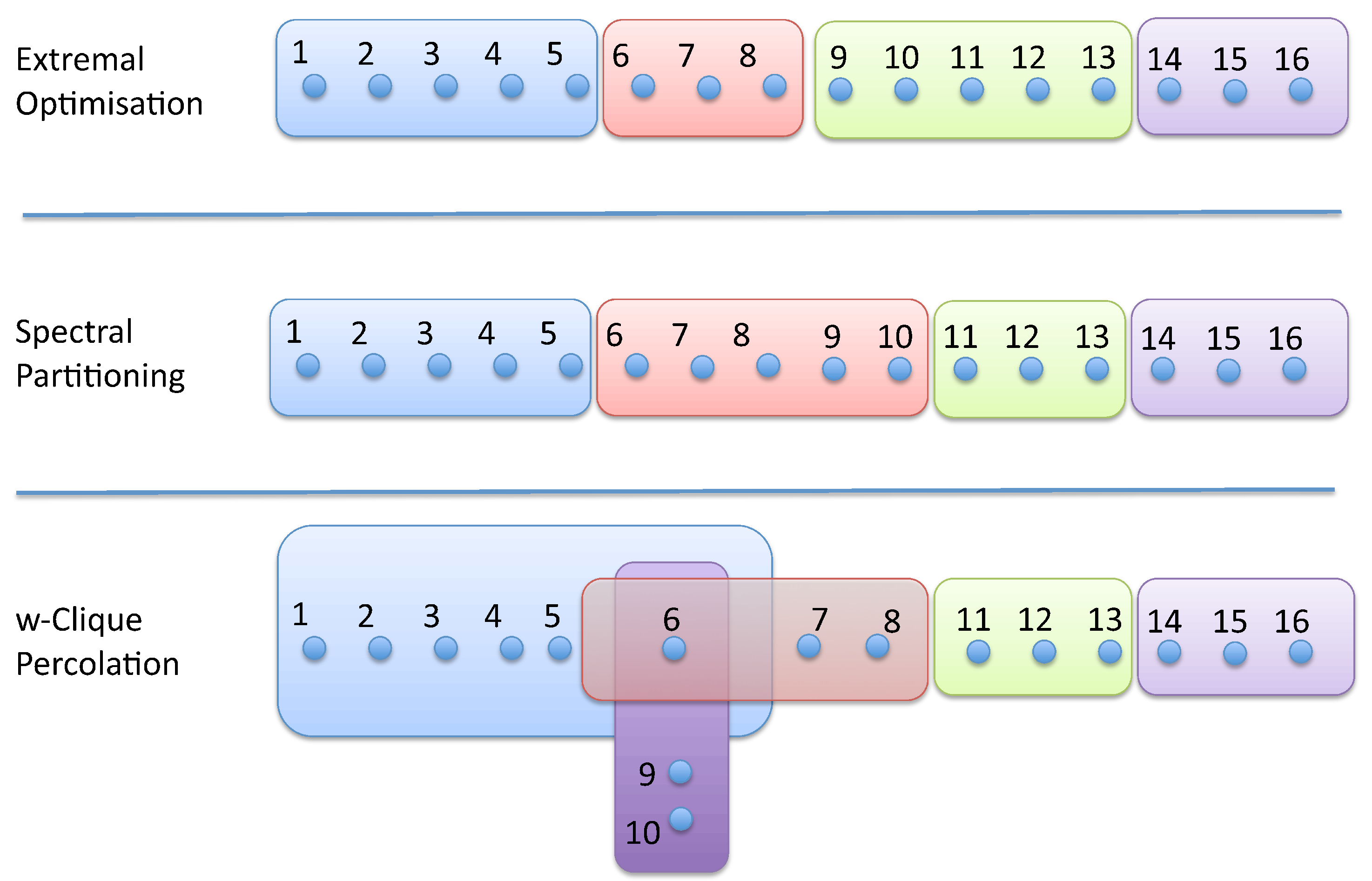}
   \caption{\label{group_memberships} Node allocations to communities for three different community detection methodologies.}
\end{figure}

Although a trivial problem at first glance, each community detection
method we applied yielded different modules and node allocations, as
seen in Fig. \ref{group_memberships}. Hard-partitioning methods such
as EO and SP produce such inconsistencies mainly due to the `broker'
nature of nodes such as $6, 9$ or $10$, which lie on high-flow paths
in the network, making them difficult to assign on one module or the
other \cite{fortunato_overview}. Although this issue is addressed by wCPM, which allows node
membership to multiple modules, it does not provide some measure of `participation strength' or `degree of belief in membership'.

\begin{figure}[htbp]
   \centering
   \includegraphics[angle=0,scale=0.6]{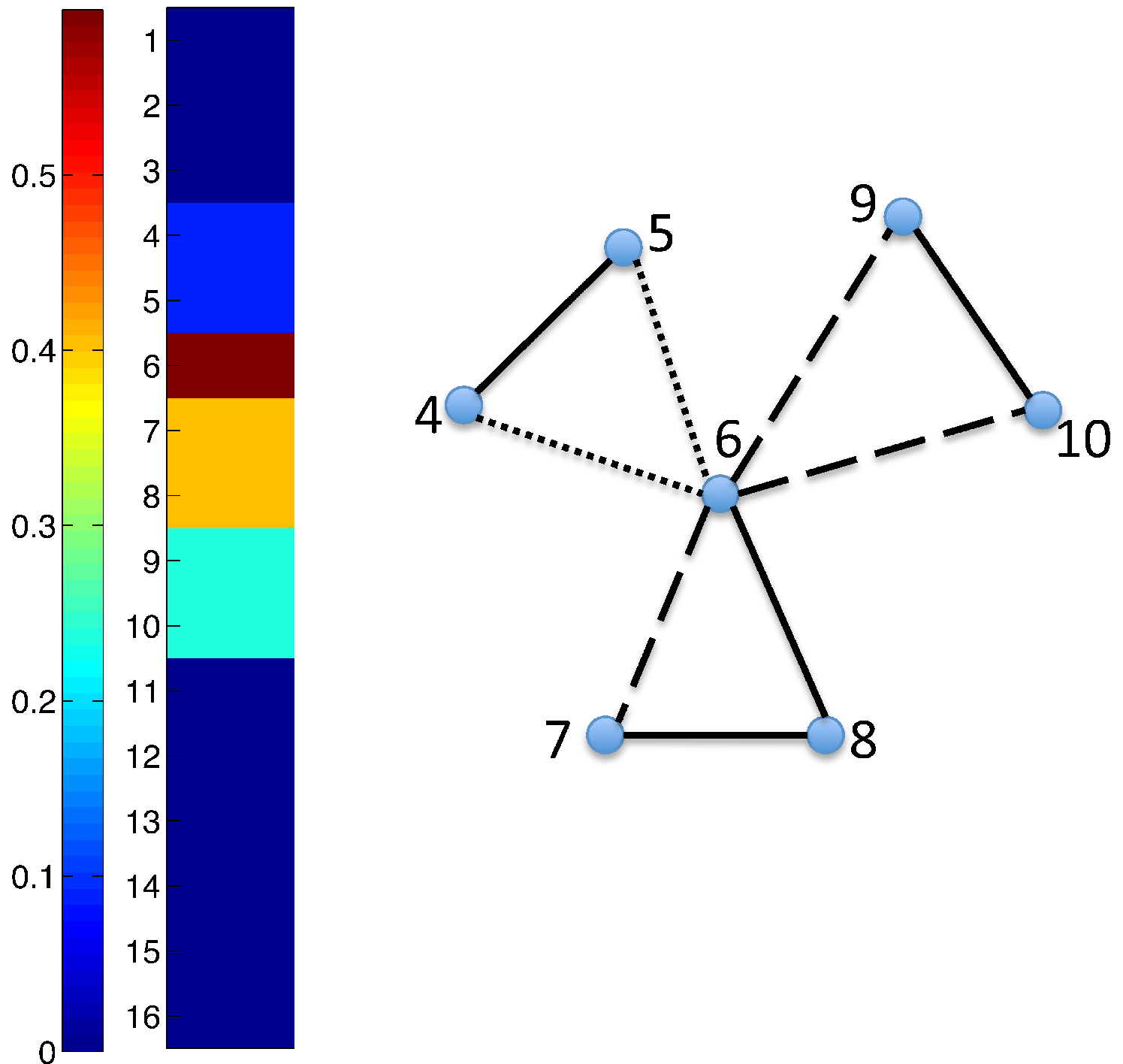}
   \caption{\label{basis} One of the NMF \emph{basis structures}, as extracted from our toy graph. Each atom has a different degree of participation, as it can be seen from the colourmap on the left. Node 6 is a focal individual, contributing the most energy to the structure along with nodes 7 and 8, while nodes 4,5,9 and 10 are peripheral.}
\end{figure}

In NMF, communities are viewed as \emph{basis structures}, captured in the model as the $K$ columns of our basis matrix $\mathbf{W}$ (see Section \ref{sec:Methods}). In this framework, we consider each basis structure or community $k$ to have a total binding energy that is allocated to the atoms (nodes) based on $\mathbf{w}_k \in \mathbb{R}^{N \times 1}_{+}$. For example in Figure \ref{basis}, we take a column of $\mathbf{W}$ and draw a colormap (left frame) based on the intensity of its elements. Components with non-zero energy correspond to nodes that participate in such basis structure and form a subset of the whole network (right frame). We can see that this basis community in Fig. \ref {basis} is dominated by nodes 6, 7 and 8, which contribute most of the binding energy, while the peripheral nodes 4, 5, 9, 10 have some minor participation. 

\begin{figure}[htbp]
   \centering
   \includegraphics[angle=0,scale=0.6]{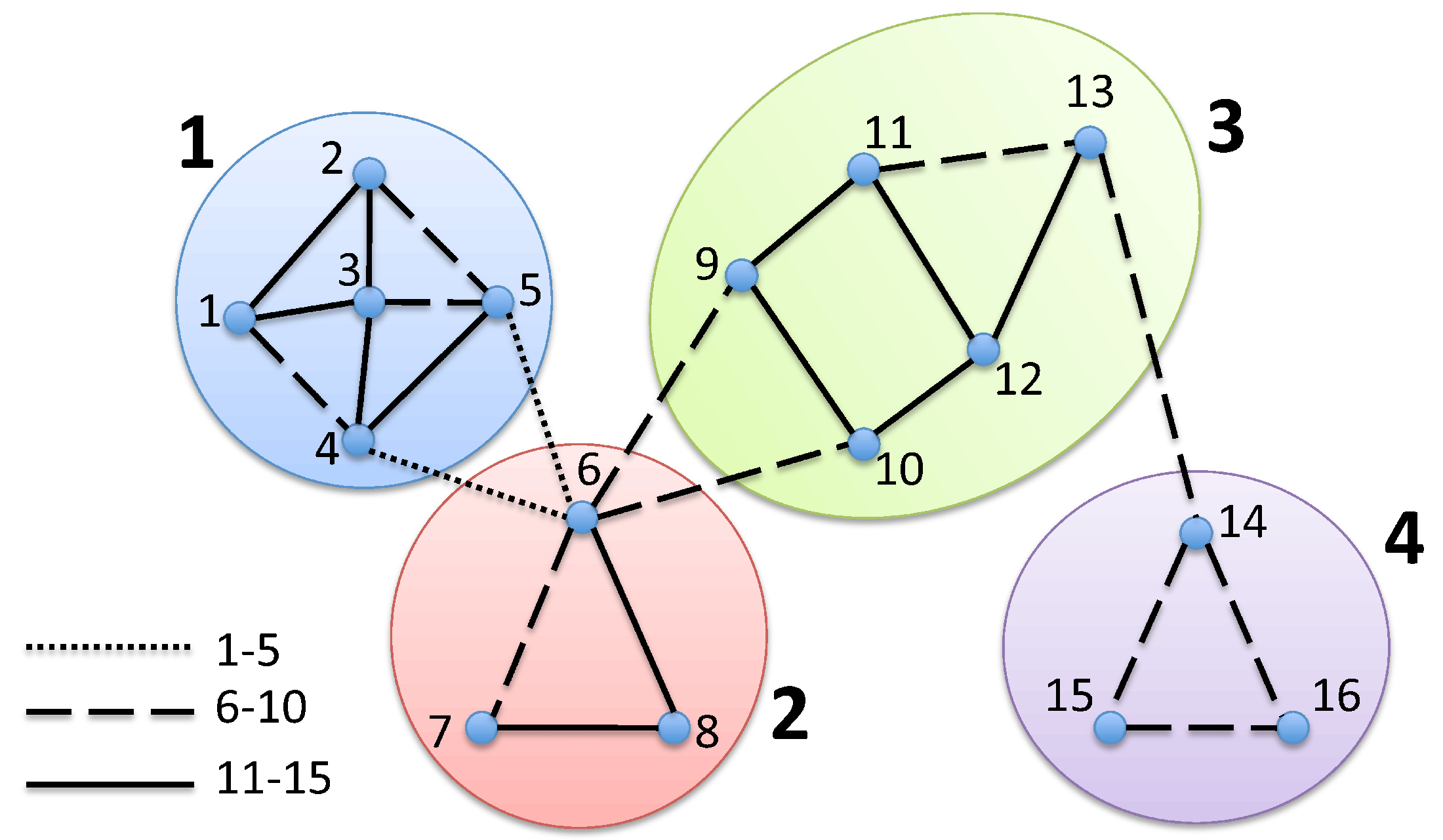}
   \caption{\label{toygraph2} The community structure of our toy graph, where each coloured plate represents a different basis structure. For purposes of illustration, we assigned each node to the community with the highest membership probability.}
\end{figure}

\begin{figure}[htbp]
   \centering
   \includegraphics[angle=0,scale=0.6]{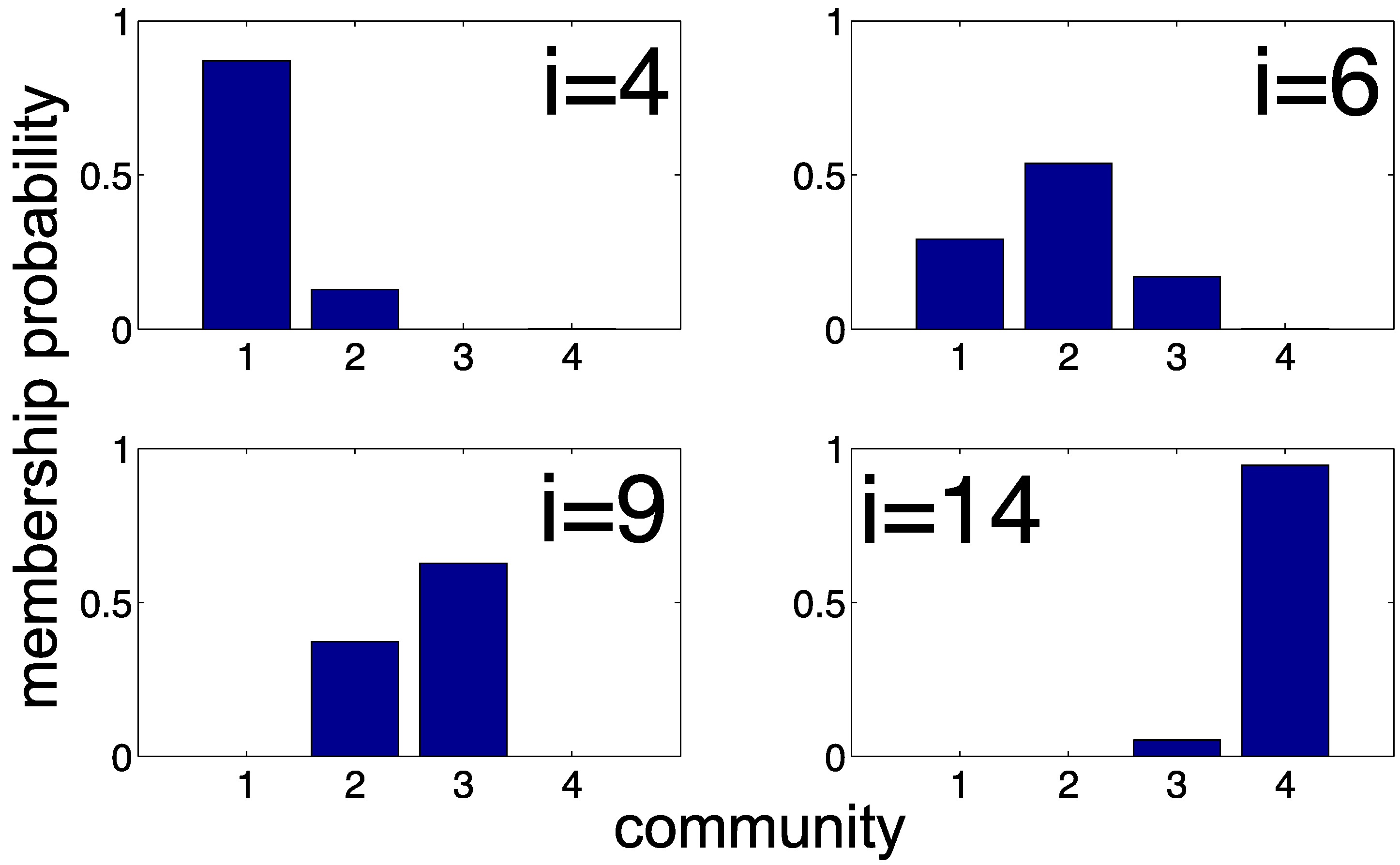}
   \caption{\label{pmembership} For each node of the toy graph our method gives a probability membership distribution; in the horizonal axis we enumerate each community appearing in Fig. \ref{toygraph2} and each bar represents the likelihood that node $i$ belongs to the respective module. For nodes that weakly communicate with other groups (such as 4 and 14) we see confident allocations, while individuals that lie on the boundary between communities (such as 6) have more entropic membership distribution.  }
\end{figure}

We applied NMF to our synthetic graph, where we extracted $K^*=4$ communities (bases to which at least one member is allocated) as seen from the four numbered plates in Figure \ref{toygraph2}. For illustrative purposes, we assigned nodes to communities using greedy allocation, i.e. we put the individual to the community into which it contributes most of its energy. The contribution of each atom $i$ to the total binding energy of each structure $k$, as denoted by $\mathbf{w}_i \in \mathbb{R}^{1 \times K}_{+}$, can be viewed as the individual's \emph{degree of participation} or, when normalised, \emph{probability of membership} to that community. From our example, in Figure \ref{pmembership} we show the different membership distributions for four different nodes in the graph. In accordance to our intuition, we see that node 6, which acts as a mediator between different communities has a more entropic membership distribution while nodes such as 4 or 14 have more confident assignments.

%while its connectivity to the other nodes (the $i$-th row or column of the adjacency matrix) is realised as a linear combination of those basis vectors.

We also can view real-world systems, such as social networks, under the framework we described above; groups of individuals are structures bound together with a given energy (time spent together, genetic relatedness, tendency for cooperation, etc). Every individual contributes to a range of communities a certain amount of such energy, which can be also seen as his/her degree of membership. High-energy members can be regarded as \emph{focal individuals} in a group, while social structures with members of uniform contribution can be regarded as \emph{teams} that are held together because of equal participation of their members. Finally, under this framework we can identify highly social individuals, that belong to many groups with high amount of participation.

Having used a simple graph to illustrate the intuition behind NMF-based community detection, we proceed in the following section to demonstrate its performance on artificial problems of larger scale and complexity.

\subsection{Benchmark datasets}

A very popular evaluation methodology for a community detection algorithm is to test it against an artificial network with ``observed'' community structure and measure how well the algorithm extracts the underlying mesoscopic organisation. The partition quality is usually compared with the original using the popular Normalised Mutual Information (NMI) criterion \cite{danon}. %, which has been recently extended \cite{overlapNMI} for the case of overlapping community structures.

\begin{figure}[h!]
\begin{center}
\includegraphics[scale=0.40]{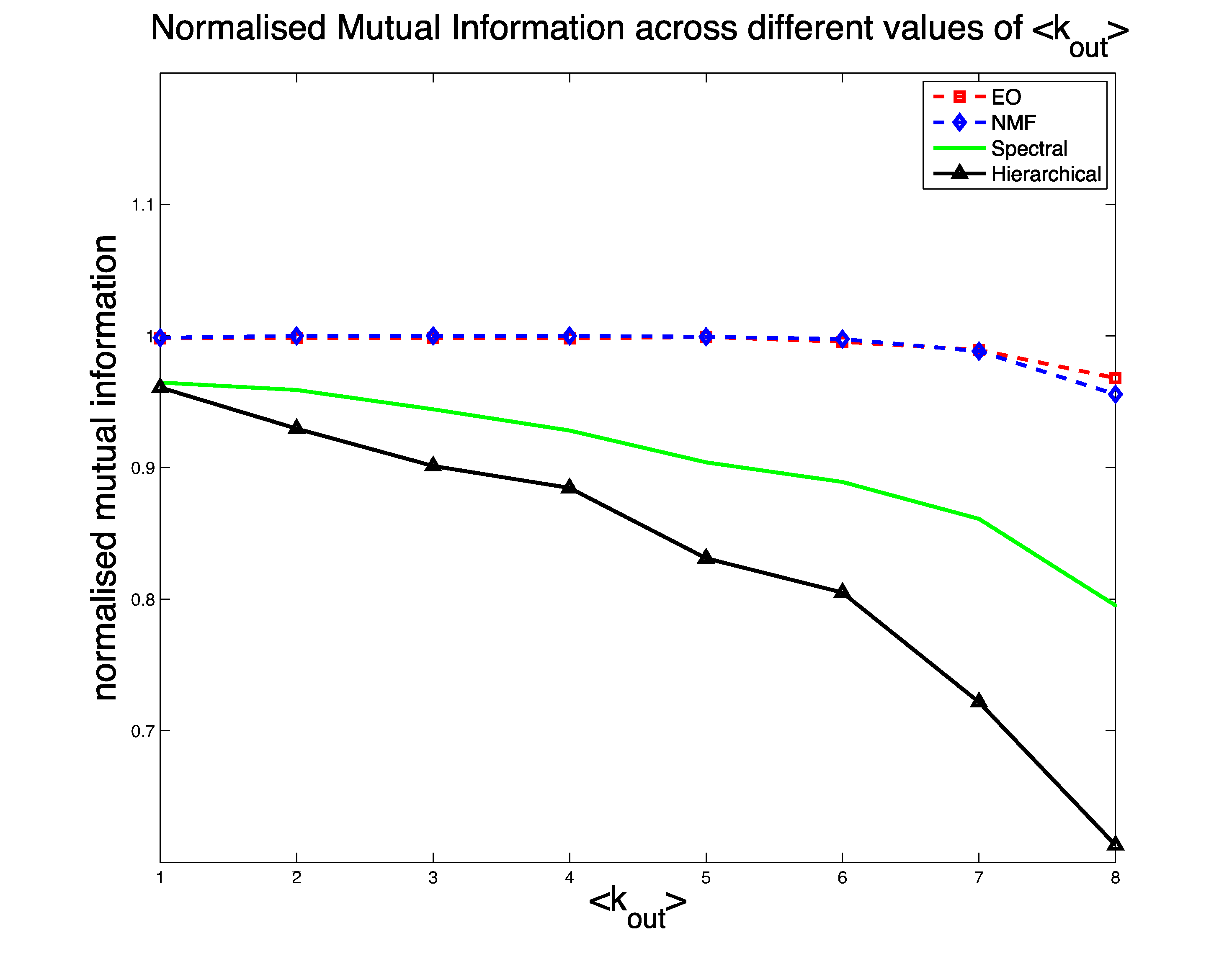}
\end{center}
\caption{\label{newman-girvan-test} Normalised Mutual Information and modularity across different levels of community cohesion in a Newman-Girvan random network. We compare our method (NMF) against other popular community detection methodologies such as Extremal Optimization (EO), Spectral Partitioning (Spectral) and Hierarchical Clustering (Hierarchical).}
\end{figure}

\begin{figure}[h!]
\begin{center}
\includegraphics[scale=0.45]{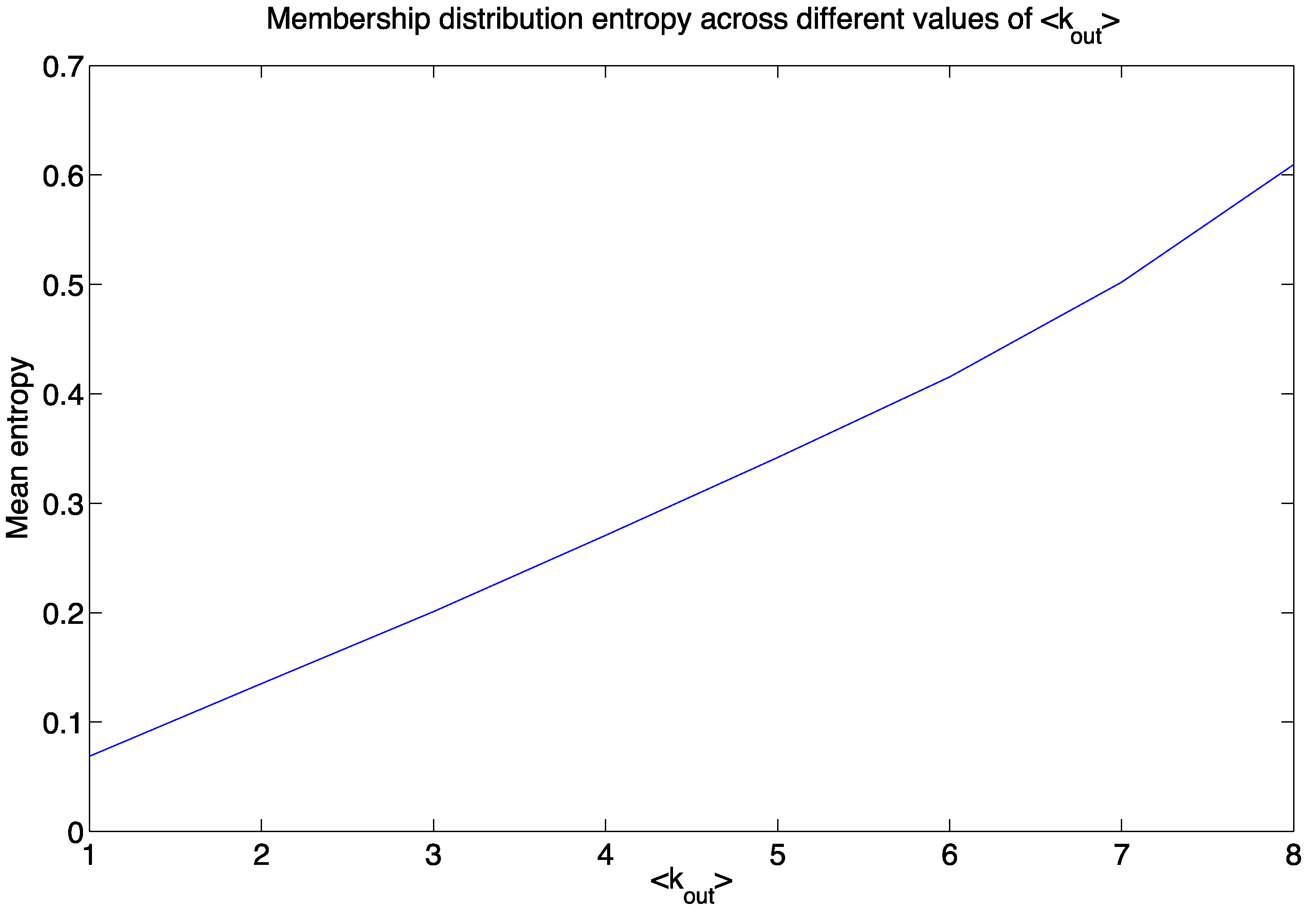}
\end{center}
\caption{\label{NG_entropy} Mean entropy (in bits) of NMF node membership probabilities for decreasing levels of community cohesion in a Newman-Girvan random graph. We notice that NMF can describe the ever-increasing fuzziness of the NG graph in terms of decreasing node allocation confidence.}
\end{figure}

We start with arguably the most popular type of benchmark problem: realisations of the Newman-Girvan random graph \cite{ng_graph} (NG graph). We generate networks with $N=128$ nodes and $C=4$ communities with $n=32$ nodes each where each one has an average degree of $\langle k \rangle=16$.  By manipulating the expected \emph{inter-community} degree $\langle k_{\rm out} \rangle$ of nodes we test our algorithm against various levels of community cohesiveness. As seen in Figures \ref{newman-girvan-test} and \ref{NG_entropy} NMF produces state-of-the-art performance in extracting the original modules for any degree of fuzziness in the artificial network, outperforming the popular Spectral Partitioning and Hierarchical Clustering (complete linkage - angular distance) method and having similar performance to Extremal Optimisation.

Although the NG graph is a very popular benchmark problem, it has been heavily criticised \cite{LaFort} for not reflecting the properties of real-world networks; NG graph realisations are small in size, with a fixed number of communities and fixed community populations while the degree distributions are uniform. For those reasons, Lancichinetti and Fortunato proposed a new class of benchmark problems \cite{LaFort}  (which we shall refer to them as LF graphs) that produce networks of any size,  with \emph{power-law} degree and community size distributions. The community cohesiveness is controlled by a \emph{mixing parameter} $\mu_t$, that signifies the expected fraction of intercommunity links per node. For the case of weighted LF graphs, we have a similar parameter $\mu_w$ that controls the strength allocation of a node between same-community members and outsiders.

\begin{figure}[h!]
\begin{center}
\includegraphics[scale=0.40]{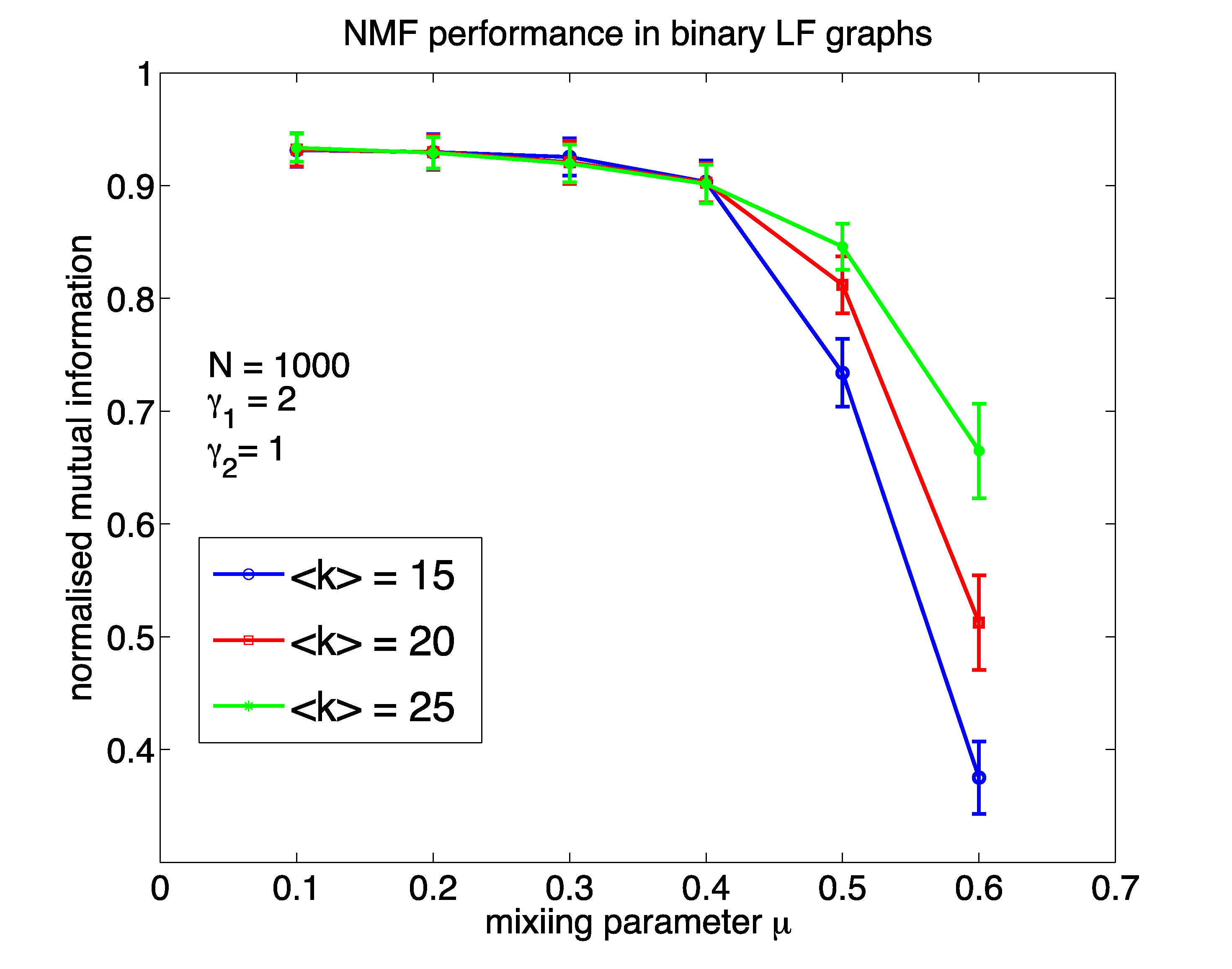}
\includegraphics[scale=0.35]{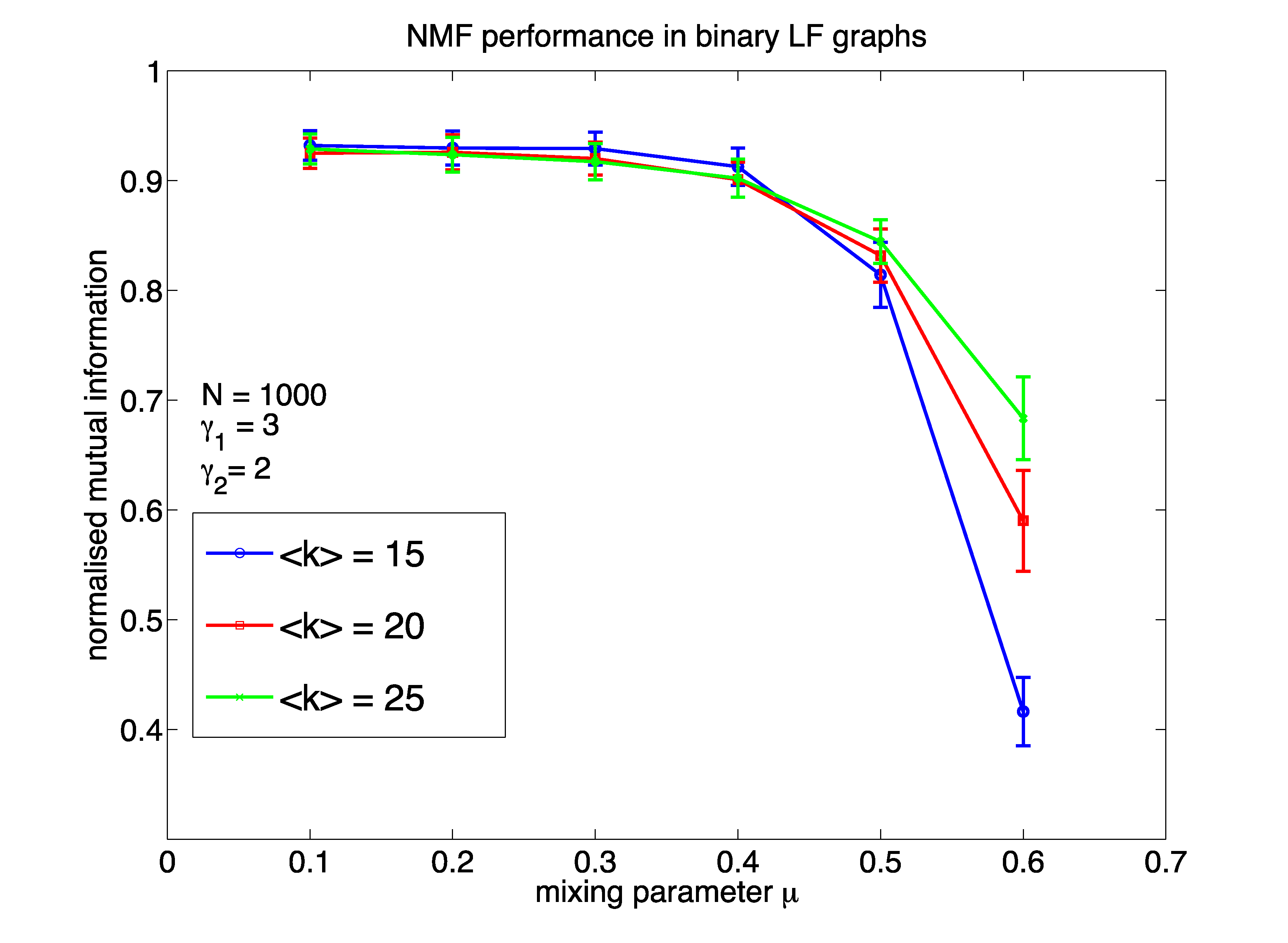}
\end{center}
\caption{\label{LFresults_binary} Normalised Mutual Information across different levels of community cohesion in binary LF random networks. We start with a very cohesive network (low mixing parameter $\mu$) and proceed by making the network fuzzier. Each point in the graph represents the average over 100 realisations of an LF graph with the given parameters. The error bars represent one standard deviation.}
\end{figure}

\begin{figure}[h!]
\begin{center}
\includegraphics[scale=0.40]{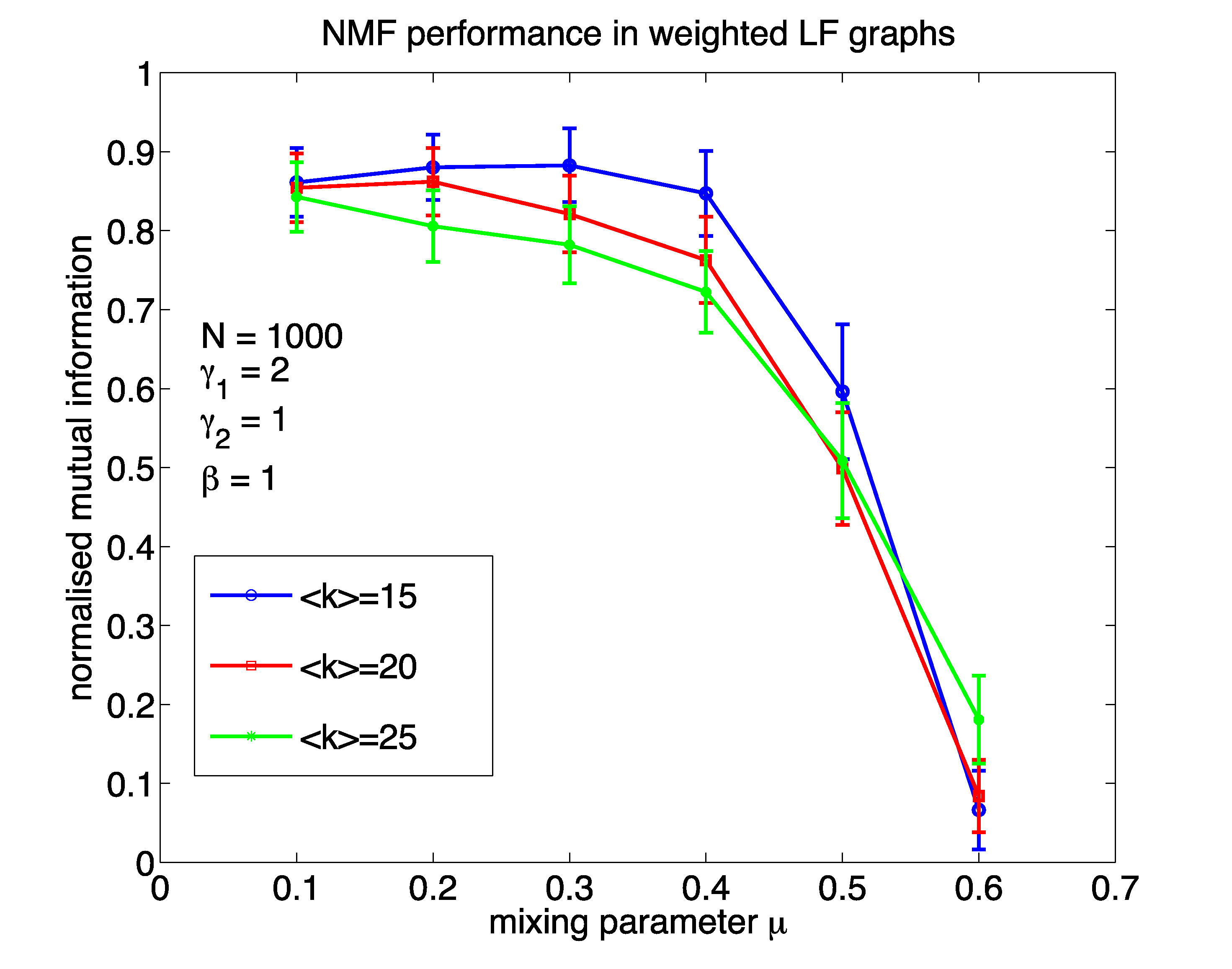}
\includegraphics[scale=0.35]{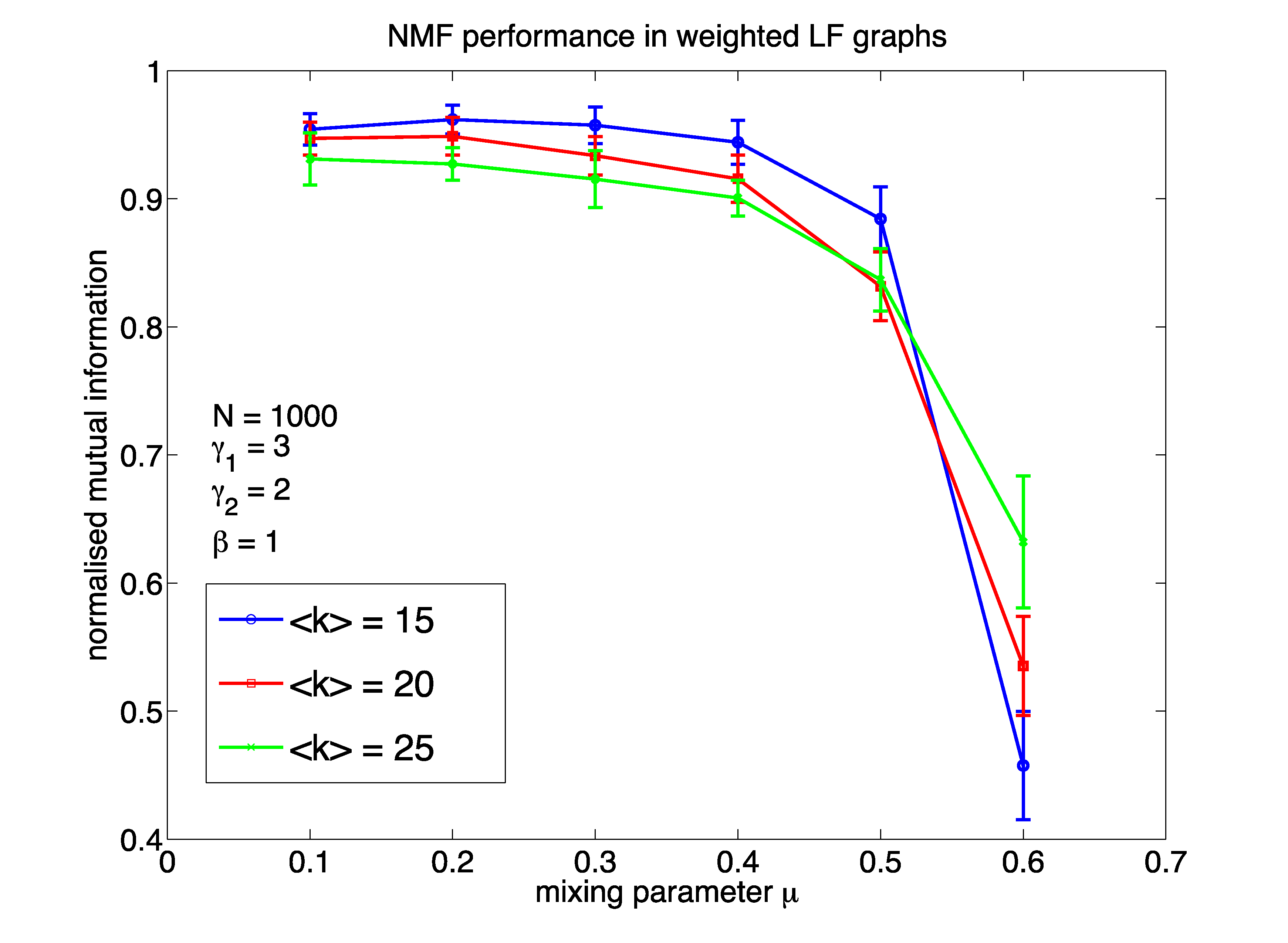}
\end{center}
\caption{\label{LFresults_weighted} Normalised Mutual Information across different levels of community cohesion in weighted LF random networks. Again, we start with a very cohesive network and proceed by making the network fuzzier. In this case of weighted network, we have the same mixing parameter for both intercommunity degrees and strengths ($\mu_t = \mu_w$). Each point in the graph represents the average over 100 realisations of an LF graph with the given parameters and the error bars represent our standard deviation.}
\end{figure}

For the purposes of our experiments, we generated a variety of such networks with $N=1000$ nodes and different parameters regarding the average degree $\langle k \rangle$ and the exponents $\gamma_1$, $\gamma_2$   of the degree and community size distributions. By starting with a small mixing parameter $\mu_t=0.1$ and for each 0.1-step up to $\mu_t=0.6$ (from `clear' to `fuzzy' community structure), we generate 100 realisations of the LF graph and monitor the module recognition performance of NMF using the popular normalised mutual information criterion. For the case of weighted LF graphs, we manipulate both mixing parameters at the same time, therefore $\mu_t = \mu_w$. The results, both for binary and weighted networks and for different configuration parameters are shown in Fig. \ref{LFresults_binary} and \ref{LFresults_weighted}.

\subsection{Real-world datasets}

In this section we present the performance of NMF on a variety of popular community detection problems. For these networks we have no ``observed solution'', therefore we measure the performance of our algorithm using the very popular Newman-Girvan \emph{modularity} $Q$ \cite{newman_girvan}. In Table \ref{tab:datasets} we present a list of our datasets, along with their number of nodes $N$ and edges $M$. Our algorithm is compared on the same data with Extremal Optimisation (EO) \cite{eo} and the Louvain \cite{louvain} methods. We note other methods such as Spectral Partitioning and Hierarchical Clustering algorithms give significantly worse performance than either NMF, EO or Louvain and these results are not presented here.

\begin{table}[ht]
\caption{Real-world datasets}
\label{tab:datasets}
\centering
\begin{tabular}{|r|r|r|r|}
\hline
Dataset&$N$&$M$&weighted? \\
\hline
\hline
Dolphins \cite{dolphins} & 62 & 159 & no \\
Books US Politics \cite{polbooks} & 105 & 441 & no \\
Les Miserables \cite{lesmis} & 77 & 254 & yes \\
College Football \cite{ng_graph} & 115 & 613 & no \\
Jazz Musicians \cite{jazz} & 198 & 2742 & no \\
C. elegans metabolic \cite{celegans} & 453 & 2025 & no \\
Network Science \cite{newman_girvan} & 1589 & 2742 & yes \\
Facebook Caltech \cite{facebook} & 769 & 16656 & no \\

%US Power Grid \cite{powergrid} & 4941 & 6594 & no \\%& 0.68 $\pm$ 0.02 & 245 $\pm$ 20.9 \\
%US Airports \cite{us_airports} & 500 & 2980 & yes \\% 0.3114 $\pm$ 0.01 & 24.33 $\pm$ 2.1 \\ 
\hline
\end{tabular}
\end{table}

For each dataset we run NMF and EO 100 times with different random initialisations and monitored the values of \emph{modularity} $Q$ along with the number $K^*$ of extracted communities. As previously detailed, for NMF initialisation, we assume a possible  maximum number of communities, $K$, equal to the number of nodes $K = N$  (which is the maximum possible partition size for any network - though we find that running with a lower value is preferred, as this reduces computation) and the `effective' number of communities $K^*$ is then inferred from the data.  The Louvain method has a very stable behaviour across different runs so we have omitted the standard deviation of modularity and community sizes for each dataset.  The algorithmic complexity of our approach is $\mathcal{O}(NK)$, as compared to $\mathcal{O}(N^2 \log N)$ for EO \cite{danon}. We also note that, in practice, as EO requires stochastic steps, the run times of the two algorihms differ even more significantly. In the majority of applications, the maximum likely number of communities $K \ll N$ and so our approach can be very efficient and competitive against the Louvain method.

\begin{table}[ht]
\caption{Modularity results against Extremal Optimisation and Louvain method}
\label{tab:r1_results}
\centering
\begin{tabular}{|r|r|r|r|}
\hline
Dataset& NMF  & EO & Louvain \\
\hline
\hline
Dolphins & 0.47 $\pm$ 0.03 & 0.51 $\pm$ 0.01 & 0.52 \\
Books US Politics & 0.52 $\pm$ $\epsilon$ & 0.48 $\pm$ 0.01 & 0.50 \\
Les Miserables & 0.53 $\pm$ 0.02 & 0.53 $\pm$ 0.01 & 0.57 \\
College Football & 0.60 $\pm$ $\epsilon$ & 0.58 $\pm$ 0.01 & 0.60 \\
Jazz Musicians & 0.43 $\pm$ 0.01 & 0.42 $\pm$ 0.01 & 0.44 \\
C. elegans metabolic & 0.36 $\pm$ 0.01 & 0.40 $\pm$ 0.09 & 0.43 \\
Network Science & 0.83 $\pm$ 0.01 & 0.86 $\pm$ 0.01 & 0.95 \\
Facebook Caltech & 0.38 $\pm$ 0.01 & 0.37 $\pm$ 0.01 & 0.37 \\

\hline
\end{tabular}
\end{table}

\begin{table}[ht]
\caption{NMF community sizes compared to Extremal Optimisation and Louvain method}
\label{tab:r2_results}
\centering
\begin{tabular}{|r|r|r|r|}
\hline
Dataset& NMF  & EO & Louvain \\
\hline
\hline
Dolphins & 6.67 $\pm$ 0.83 & 4 $\pm$ 0 & 5 \\
Books US Politics & 6.23 $\pm$ 0.62 & 4.04 $\pm$ 0.4 & 3 \\
Les Miserables & 9.97 $\pm$ 0.78 & 4.96 $\pm$ 1.72 & 6 \\
College Football & 8.86 $\pm$ 0.79 & 8 $\pm$ 0 & 10 \\
Jazz Musicians & 8.57 $\pm$ 8.89 & 4 $\pm$ 0 & 4 \\
C. elegans metabolic & 15.69 $\pm$ 1.14 & 7.96 $\pm$ 1.06 & 10 \\
Network Science & 342.53 $\pm$ 5.28 & 58.24 $\pm$ 12.36 & 418 \\
Facebook Caltech & 24.28 $\pm$ 1.72 & 6.84 $\pm$ 1.82 & 10 \\

\hline
\end{tabular}
\end{table}

In Table \ref{tab:r1_results} we present our experimental results for each dataset of Table \ref{tab:datasets}. We use the popular Newman-Girvan modularity $Q$ as a performance measure of partition quality and we also present the number of identified communities. As $Q$ can not account for the overlapping nature of communities, we use `greedy allocation' i.e we assign a node to the module with the highest probability of membership. The results of NMF are presented alongside the very popular Extremal Optimisation  and Louvain method for comparative analysis. From Table \ref{tab:r1_results} we can see that our approach performs competitively yet is not an algorithm designed with the aim of maximising modularity, unlike either EO or the Louvain methods. Additionally, it has the advantage of providing \emph{probabilistic outputs} for community membership (therefore achieving soft partitioning) and having \emph{low computational overhead}. Finally, NMF does not suffer from the resolution limit \cite{resolution} of modularity optimisation methods such as EO, where smaller groups are merged together \cite{mason_overview} \cite{resolution}, ending up with smaller number of communities, as seen in Table \ref{tab:r2_results}.

%%%%%%%%%%%%%%%%%%%%%%%%%%%%%%%%%%%%%%%%%%%%%%%%%%%%%%%%%%%%%%%%%%%%%%%%%%

\section{Conclusions} 
\label{sec:Conclusion}

In this work we described a novel approach to community detection that adopts the Bayesian non-negative matrix factorisation model of \cite{tan09} to achieve soft-partitioning of the network, assigning each node a probability of membership over all the extracted communities. That allows us not only to capture the fuzziness of the network (via the entropy of the membership distribution) but also to improve network cartography techniques \cite{guimera} by identifying central and peripheral nodes in modules. Network visualization tools can also be improved in this manner. The approach is computationally efficient and offers performance comparable to state-of-the-art methods. Indeed the performance advantages for large data sets allow the NMF approach to be run many times compared to a single run of competing approaches. Clearly this allows for the selection of the best performing run, or a small ensemble of high-modularity solutions.

\section{Future work} \label{sec:future_work}
Future application work in this area addresses the analysis of a large zoological data set of interactions between members of a population of wild birds. As significant data exists and secondary verification data has been collated, such as breeding pair identifications etc. this offers a unique chance to verify any relationships that our approach detects.

Work is currently underway to allow this model to be extended to incorporate dynamics such that non-stationary, time-varying, community relationships may be tracked. We have not discussed the handling of missing data in this paper, but taking missing observations into account may be readily handled in our approach and this is detailed in \cite{Cemgil:09}. As mentioned in the paper, more complex priors over the set of $\beta_k$ would allow for domain knowledge to be incorporated and for correlated structures to be correctly dealt with.

\section{Acknowledgements} \label{sec:Acknowledgements}
The authors would like to thank Nick Jones, Mason Porter and Mark Ebden for valuable comments. Ioannis Psorakis is funded from a grant via Microsoft Research, for which we are most grateful.

\bibliographystyle{plain}
\bibliography{commDetNMF.bib}

\end{document}